\documentclass[sigconf,authorversion,nonacm]{acmart}

\setcopyright{acmcopyright}
\copyrightyear{2021}
\acmYear{2021}
\acmDOI{10.1145/1122445.1122456}

\usepackage{multirow}

\def\rot{\rotatebox}
\begin{document}

\title{CalCROP21: A Georeferenced multi-spectral dataset of Satellite Imagery and Crop Labels}

\author{Rahul Ghosh}
\authornote{Both authors contributed equally to this research.}
\email{ghosh128@umn.edu}
\affiliation{%
  \institution{University of Minnesota}
  \city{Minneapolis}
  \state{MN}
  \country{USA}
}

\author{Praveen Ravirathinam}
\authornotemark[1]
\email{pravirat@umn.edu}
\affiliation{%
  \institution{University of Minnesota}
  \city{Minneapolis}
  \state{MN}
  \country{USA}
}

\author{Xiaowei Jia}
\email{xiaowei@pitt.edu}
\affiliation{%
  \institution{University of Pittsburgh}
  \city{Pittsburgh}
  \state{PA}
  \country{USA}
}

\author{Ankush Khandelwal}
\email{khand035@umn.edu}
\affiliation{%
  \institution{University of Minnesota}
  \city{Minneapolis}
  \state{MN}
  \country{USA}
}

\author{David Mulla}
\email{mulla003@umn.edu}
\affiliation{%
  \institution{University of Minnesota}
  \city{St. Paul}
  \state{MN}
  \country{USA}
}

\author{Vipin Kumar}
\email{kumar001@umn.edu}
\affiliation{%
  \institution{University of Minnesota}
  \city{Minneapolis}
  \state{MN}
  \country{USA}
}

\renewcommand{\shortauthors}{Ghosh, et al.}

\begin{abstract}
Mapping and monitoring crops is a key step towards sustainable intensification of agriculture and addressing global food security. A dataset like ImageNet that revolutionized computer vision applications can accelerate  development of novel crop mapping techniques. Currently, the United States Department of Agriculture (USDA) annually releases the Cropland Data Layer (CDL) which contains crop labels at 30m resolution for the entire United States of America. While CDL is state of the art and is widely used for a number of agricultural applications, it has a number of limitations (e.g., pixelated errors, labels carried over from previous errors and absence of input imagery along with class labels). In this work, we create a new semantic segmentation benchmark dataset, which we call CalCROP21, for the diverse crops in the Central Valley region of California at 10m spatial resolution using a Google Earth Engine based robust image processing pipeline and a novel attention based spatio-temporal semantic segmentation algorithm STATT. STATT uses re-sampled (interpolated) CDL labels for training, but is able to generate a better prediction than CDL by leveraging spatial and temporal patterns in Sentinel2 multi-spectral image series to effectively capture phenologic differences amongst crops and uses attention to reduce the impact of clouds and other atmospheric disturbances. We also present a comprehensive evaluation to show that STATT has significantly better results when compared to the resampled CDL labels. We have released the dataset and the processing pipeline code for generating the benchmark dataset.
\end{abstract}

\maketitle

\section{Introduction}
With the rise in world's population, food supplies must scale up to keep pace with the growing demand. Hence it is critical to ensure that farm lands are being used efficiently from an environmental perspective.
In particular, mapping and monitoring crops is a key step towards forecasting yield, guiding sustainable management practices, measuring the loss of productive cropland due to urbanization and evaluating progress in conservation efforts.

In the United States, the Department of Agriculture’s (USDA) Cropland Data Layer (CDL) provides a publicly available land-cover classification map annually at 30m resolution which includes major crop commodities for the conterminous United States (CONUS)~\cite{CDL}. CDL product has driven the advancement of research in areas ranging from agricultural sustainability studies~\cite{hartz2011greenbrier,fitzgerald2013agriculture}, to environmental issues~\cite{belden2012assessment,cibin2012simulated}, land conversion assessments~\cite{wright2013recent,rashford2013modeling}, crop rotations~\cite{plourde2013evidence,boryan2012deriving}, farmer surveys~\cite{painter2013results} and many more~\cite{cdluse}. While CDL is the state-of-the-art spatially explicit identification product for crops, it has a number of limitations~\cite{reitsma2015does,pritsolascautionary}. First, the CDL is created using a pixel based classification algorithm and hence contains pixelated errors in crop labels. Second, each pixel is not updated every year and labels for some pixels are borrowed from previous years which sometimes leads to incorrect labels. Third, CDL is known to have low accuracy in classifying many minor crops such as alfalfa, hay, tree crops, and many vegetable crops \cite{accbias2021}. Finally, CDL labels are created using Landsat images, which are at 30m resolution, leading to mixed pixels errors. The Sentinel constellation provides images at a finer resolution (10m) and more frequent temporal scale (5days vs 15 days) and thus offers the possibility of creating crop labels at 10m resolution.

The success of deep learning in solving highly complex tasks in the field of computer vision  and natural language processing can be attributed to the availability of large datasets such as ImageNet~\cite{ImageNet} and computational resources. These large datasets are essential for the generalization of such deep learning methods. Indeed advances in Earth observation technologies have led to the collection of vast amount of accurate and reliable remote sensing (RS) data that provide tremendous potential to create similar large scale datasets for mapping crops over large regions. For example, several large-scale remote sensing datasets have been created for similar land cover mapping tasks~\cite{UCMerced,DeepSat,BrazilCoffee,eurosat,Bigearthnet,Inria,DeepGlobe,sen12ms}. However, these datasets cannot be directly used for crop mapping for several reasons. First, existing large-scale RS datasets mostly provide a single view of the earth's surface in time. As highlighted in previous literature~\cite{ghosh2021attentionaugmented}, the distribution and growth of crops commonly exhibit special spatio-temporal patterns, e.g., the contiguous nature of crop fields and temporal/seasonal patterns due to  their unique crop phenology. Hence, these datasets are not designed for effective crop mapping using spatial and temporal data patterns. Moreover, these large scale datasets are either limited ~\cite{ghosh2021land} in the number of categories (i.e limited number of classes that do not cover many important crop types)~\cite{DeepGlobe} or lack in the spatial resolution (e.g. single label assigned to entire image)~\cite{Bigearthnet}.

To overcome the limitations of existing RS datasets and to facilitate deep learning research in RS-based crop mapping, this paper presents a new semantic segmentation benchmark dataset for crops, CalCROP21. Specifically, the input images were created using a Google Earth Engine based robust image processing pipeline on the multi-spectral temporal images collected by the Sentinel-2 constellation in the Central Valley of California in 2018. A novel spatio-temporal semantic segmentation \cite{ghosh2021attentionaugmented} method was used to generate better quality labels using resampled CDL as initial labels. This efficacy of the methodology relies on several key assumptions. First, the noisy coarse resolution CDL labels are still of good enough quality to be used for training a classifier. Second, a classifier that makes use of space and time is more effective in dealing with the training label noise than one that ignores such information. Third, labels at the geographical farm boundaries can be mixed and their labels at the coarse resolution are not trustworthy, whereas labels at the interior of a region are likely to be more confident.

To summarize, our contributions in this paper are as follows:
\begin{itemize}
    \item This dataset is a first large scale semantic segmentation dataset that includes both input images as well as labels for a diverse array of crops at 10m resolution. Specifically, each pixel in the image is labeled as one of 21 crop or 7 other classes.
    \item We improve the quality of the resampled CDL labels for these classes using a novel spatio-temporal deep learning method based on the phenotypic differences among crops with Sentinel images at multiple time steps.
    \item We validate the quality of the labels via a detailed quantitative and qualitative evaluation.
    \item We provide the processing pipeline code for further use by the community in collecting images and generating results for a different year and using different temporal frequency.
    \item With some small edits, this pipeline can be used to produce similar results for any other part of the US or the world (assuming some initial labels are available even at a coarse resolution).
\end{itemize}

\section{Related Works}
Several such benchmark datasets are available for land use and land cover (LULC) mapping. They  can be divided into multiple categories based on their downstream tasks, a) object recognition \cite{DOTA}, b) image classification \cite{UCMerced,DeepSat,BrazilCoffee,eurosat,Bigearthnet} and c) semantic segmentation \cite{Inria,DeepGlobe,sen12ms}. These datasets do not have sufficient granularity to enable scientific advances in crop mapping using deep learning. First, they provide labels that cover a very small number of categories. Second, due to the use of overhead aerial imagery which only captures limited number of bands as compared to satellites which can capture a vast array of bands and thus help in distinguishing between crops. Unlike image classification task which only provides presence of a crop in an image, image segmentation datasets are more relevant from the domain perspective because knowledge of area under different crops provides insights into food supply. While datasets are available for LULC, to the best of our knowledge, no dataset on crop semantic segmentation that includes minor crops is available. For example, in the context of cropland mapping, BigEarthNet~\cite{Bigearthnet} is the one most relevant dataset which provides 590,326 image patches from 125 Sentinel-2 tiles and associate each image patch with a subset of 43 Corine Land Cover classes over Europe. However, there are several limitations in this dataset which make it less ideal for crop monitoring. First, the topology of the classes included do not distinguish between different crops but rather between broad vegetation types (forest, agriculture, grassland, etc). Second, due to the association of labels to entire image patch it captures the presence and not the area of a certain category of land cover. Finally, a single temporal snapshot for an area does not allow identification of different types of crops~\cite{ghosh2021attentionaugmented}.

\section{Data Sources}
We use freely available multi-spectral satellite images and data products to create our dataset. Specifically, we use Sentinel-2 as the input images and the Cropland Data Layer as initial labels. Here we describe the data sources involved in creating the dataset.

\subsection{Input Satellite Imagery}
In our dataset, we use the multi-spectral images captured by the two polar-orbiting satellites as part of the Sentinel-2 mission operated by the European Space Agency (ESA). Due to Its high revisit time of 5 days, phenological characteristics of different crops can be observed  compared to using single snapshot (or few snapshots) for the whole season. The multi-spectral images has 13 bands in the visible, near infrared, and short wave infrared part of the spectrum, each having a spatial resolution of 10, 20 or 60m. The captured images are available in the form of tiles, each of which have a unique ID and covers an area of 10,000 sq km.

\subsection{Crop Labels}
The Cropland Data Layer (CDL) is an annual publicly available land cover classification map for the entire US. CDL is generated by a Decision tree based approach using moderate resolution satellite imagery and extensive agricultural ground truth. With over 200 classes, CDL provides land cover maps covering the entire conterminous United States (CONUS) at 30-meter spatial resolution with a high accuracy up to 95\% for classifying major crop types (i.e., Corn, Soybean, and Wheat). The CDL data products are free to download from Google Earth Engine~\cite{cdl_google}. Although CDL is a very useful product that has led to the development of many downstream applications, the product is plagued with noise that arise due to the reasons discussed in Section 1. In particular, it has high accuracy (up to 95\%) for classifying major crops(e.g., Corn, Soybean, Wheat), but it is known to have poor accuracy for minor crops (e.g., Alfalfa, Hay and Tree crops) \cite{accbias2021}.

\begin{figure}[t]
    \centering
    \includegraphics[width = 0.8\columnwidth]{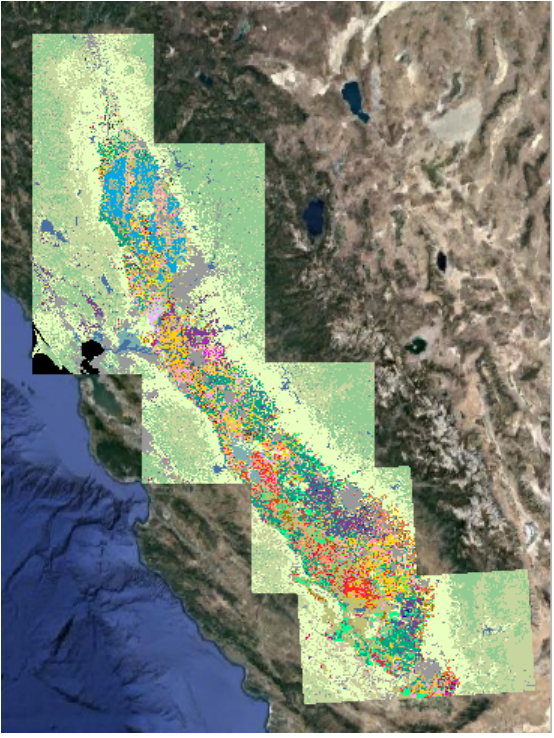}
    \caption{Overlay of 11 tiles in Central Valley of California region we used in our study. The colors represent different crops and the mapping (refer to Figure~\ref{fig:argmaxvsrapt}) is kept consistent throughout this paper.}
    \label{fig:raw_label_california_full}
\end{figure}

\section{Processing Pipeline}
We use Google Earth Engine to build a robust image processing pipeline to create biweekly Sentinel image composites. Using the obtained biweekly composites and CDL labels, we develop a novel spatio-temporal deep learning  method to improve upon the original CDL labels. In the following, we describe these steps in details.

\subsection{Generation of bi-weekly Sentinel-2 multispectral composites}
Many land covers, e.g., different types of crops, are indistinguishable at a single time step. In particular, different crops have different seeding time and harvesting time, which is also affected by  weather conditions. Hence, different crops show discriminative signatures at different points of time~\cite{ghosh2021attentionaugmented}. Correctly identifying crops will require modelling its entire agricultural cycle from sowing to harvesting. Hence, we consider all the images available in a year for this dataset. However, these images often have clouds and other atmospheric disturbances. Here, we generate bi-weekly image composites using a robust Google Earth Engine based pipeline to reduce the impact of these atmospheric disturbances. Specifically, we collect all available images within a 2-week period and score every pixel of each tile using a the quality band (QA60), which presents information as to whether a pixel is cloud-free, dense cloud or cirrus cloud. The quality scores are then thresholded to create cloud masks for each image. Based on the amount of cloud-free pixels the collection of images are sorted and finally the best images merged to create a cloud-free mosaic. Thus we obtain 24 mosaics in a year per tile, each in the projection of the zone (for California crop belt, 8 tiles are in EPSG:32610 - WGS 84 / UTM zone 10N and 3 are in EPSG:32611 - WGS 84 / UTM zone 11N). All the bands are resampled to  10m spatial resolution and then exported from Google Earth Engine. Since some Sentinel tile images may be slightly larger than the area they are supposed to cover,  we use GDAL to clip the images and reproject them so that every pixel is of 10m$\times$10m resolution. 

This finally produces 24 georeferenced files each of which has a shape of  (10980,10980,10) for a tile, with 10 signifying the number of bands (10m and 20m) used. Since our objective is to map the entire crop belt in the Central Valley of California, we found that 11 Sentinel-2 tiles covers this crop belt, namely T10SEH, T10SEJ, T10SFG, T10SFH, T10SFJ, T10SGF, T10SGG, T10TEK, T11SKA, T11SKV, and T11SLV, giving a total of 264 tif files. As a preprocessing step we first clip the bottom and top 2\%ile of each channel of the satellite images and then apply max-min normalization. Following the preprocessing of the images, we split each tile into 100 grids each of size ~10km$\times$10km (1098$\times$1098 pixels). We combine all the 24 composite images corresponding to same grid together to form an array of shape (24,1098,1098,10): 24 timestamps, (1098,1098) pixels and 10 channels. We have 1,100 grid arrays in total, each of which is named as ``TILEID\_YEAR\_ROW\_COL\_IMAGE.npy'', e.g., ``T11SKA\_2018\_5\_6\_IMAGE.npy'' corresponds to the 5th row and 6th column (indexed from 0) of the tile T11SKA in 2018.

\begin{figure}[t]
    \centering
    \includegraphics[width = \columnwidth]{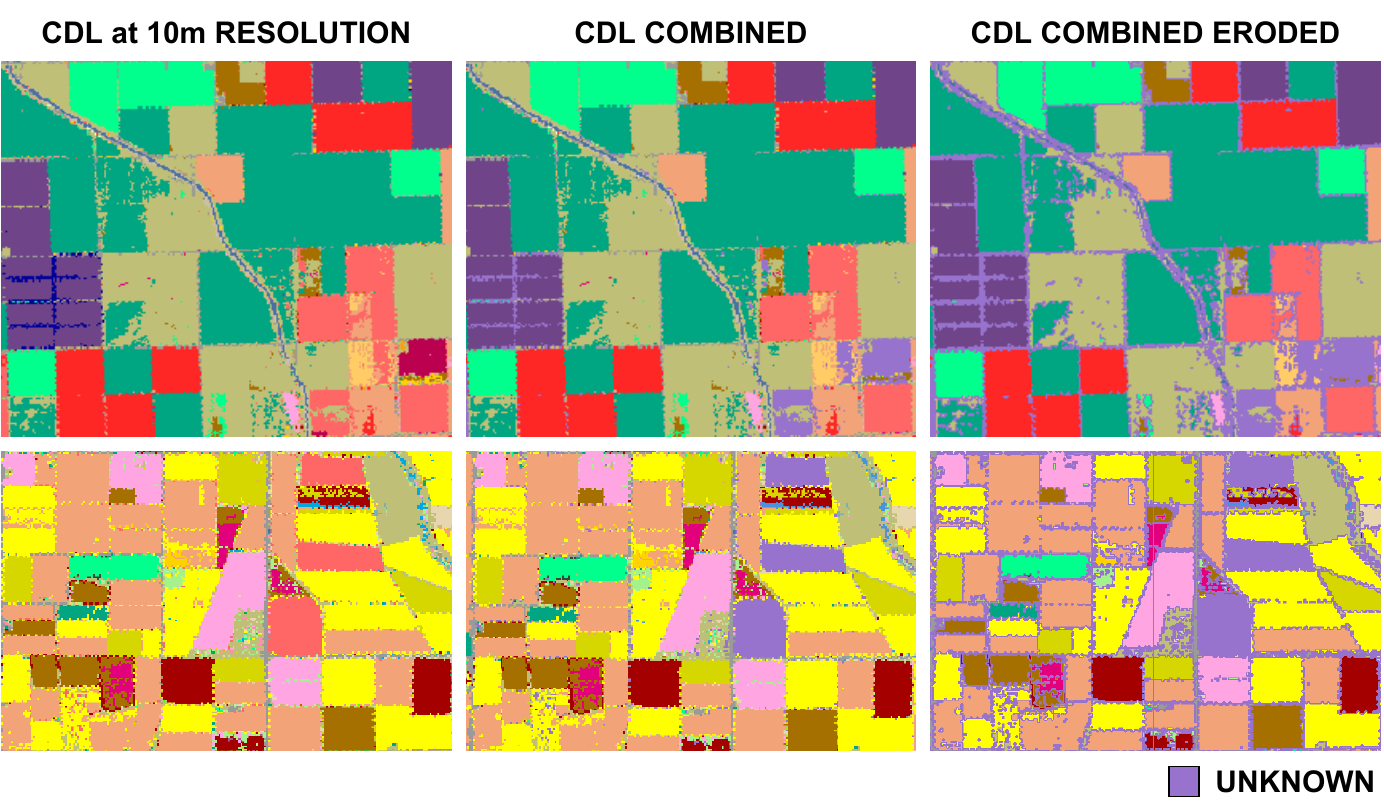}
    \caption{Evolution of labels through each preprocessing step are shown for two randomly chosen regions (rows). The first columns shows the raw CDL labels resampled at 10m resolution. The second columns shows a revised set of labels, where similar classes have been combined and untrustworthy classes have been grouped as unknown class (in light purple). The third column shows one level of erosion done classwise at boundaries and also removal of connected components of size less than 4 pixels. We refer to these labels as CDL-combined-eroded, and use these for training our deep learning model.}
    \label{fig:label_evolution}
\end{figure}

\subsection{Pre-processing of CDL}
We use Google Earth Engine to fetch the labels and crop them using each georeferenced Sentinel-2 tile, which produces a label image at 30m resolution for each  tile. We then resample the labels to 10m resolution to create 11 label tiles of shape (10980,10980). The overlay of the 11 labels tiles on the California Central Valley crop belt is shown in Fig.~\ref{fig:raw_label_california_full}. CDL provides labels for more than 200 crop classes, many of which are completely absent or rarely present in the California Central Valley region. In our dataset we exclude these absent classes in the California Central Valley region. In addition, CDL  provides state-wise validation metrics for their labels using ground-truth labels. We also exclude those classes for which the number of pixels used for CDL validation is too few as their labels cannot be trusted. Specifically, we include a crop class in our dataset if it fulfils the following conditions:
\begin{itemize}
    \item The crop class has at least 1 million pixels in the study region (in this case the 11 tiles).
    \item The crop class  has at least 100 validation pixels used by CDL.
\end{itemize}
For non-crop classes we only apply the first condition (e.g., wetlands, grass, forests, hay, urban etc.) as their validation metrics are not provided by CDL.Following these steps we are left with 34 classes: \{Corn, Cotton, Rice, Sunflower, Barley, Winter Wheat, Safflower, Dry Beans, Onions, Tomatoes, Cherries, Grapes, Citrus, Almonds, Walnut, Pistachio, Garlic, Olives, Pomegranates, Alfalfa, Hay, Barren, Fallow and Idle, Deciduous Forests, Evergreen forest, Mixed Forests, Clover and wildflower, Shrubland, Grass, Woody wetlands, Herbaceous Wetlands, Water, Urban,  Double Crops\}. For training and evaluation purposes we combine the different forest classes to a super class ``Forest Combined'', wetland classes to ``Wetlands Combined'', and  combine \{Grass, Shrubland, Clover, Wildflower\} to ``Grass combined''. We also do not use Double Crops in our study and label all those pixels as unknown class. Following the preprocessing of the labels we are left with 21 crop classes and 7 other classes and we refer to this label set as CDL-combined.

Since the CDL is originally at 30m resolution, which we resample to 10m (to match with the resolution of input images), the boundary pixels are mixed and thus they could contain regions of multiple classes. Given the uncertainty of labels at spatial boundaries between any two classes, we perform 1 pixel erosion for each class and replace these eroded pixels with unknown class and remove connected components of a class that are less than or equal to size 4. These labels are called CDL-combined-eroded. Fig~\ref{fig:label_evolution} shows the progression of the labels through these preprocessing steps. Similar to the image data, post erosion we segment and store the label in arrays of shape (1098,1098) and have the naming convention as TILEID\_YEAR\_ROW\_COL\_PREPROCESSED\_CDL\_LABEL.npy.

\subsection{Grid curation}
\begin{figure}[t]
    \centering
    \includegraphics[width = 0.8\columnwidth]{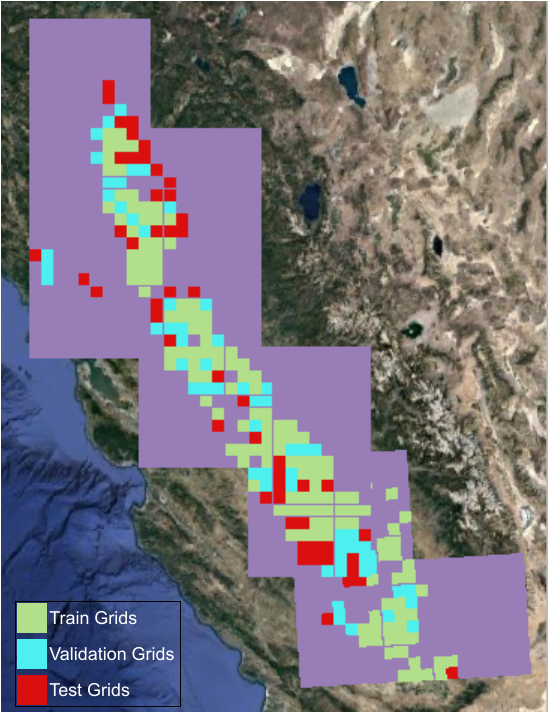}
    \caption{Distribution of Train/Val/Test grids. Green, light blue, and red represent the regions used for training, validation and testing, respectively. The purple regions denote the non-agricultural land and are not used in our experiments.}
    \label{fig:distri2}
\end{figure}

As described earlier, in our dataset we have 1,100 grids of 1098$\times$1098 pixels in size covering the entire crop belt in California's Central Valley. Many of these grids are predominantly covered by non-crop classes, and hence are removed from the dataset resulting in 367 acceptable grids. Specifically, a grid is included if it follows both of the following conditions:
\begin{itemize}
    \item Grid has at least 50\% pixels that are not unknown
    \item Out of the valid pixels, Grid G has at least 50\% pixels that belong to crop classes
\end{itemize}

\subsection{Label Improvement using STATT}
As described earlier, CDL based labels cannot be used directly as reference labels. To improve the quality of CDL labels, we used the STATT model proposed by the authors in \cite{ghosh2021attentionaugmented} which uses spatial as well as temporal information to effectively model the phenology of crops and reduce the effects of clouds and other atmospheric disturbances. Specifically STATT uses a UNET style architecture to extract spatial features and a bidirectional Long-Short Term Memory (biLSTM) to model temporal progression of the crop specific growing and harvesting patterns. Further it uses attention networks to aggregate the hidden representations for each time-step based on their contribution to the classification performance. Finally, using these attention scores, the spatial features by the convolutional encoders at multiple resolutions are aggregated and passed using skip connections to the convolutional decoder to generate segmentation maps. A comparison of STATT with alternative approaches that model either the spatial or temporal information, or both (but not as effectively as STATT) is available in \cite{ghosh2021attentionaugmented}.

To demonstrate the efficacy of this method in improving the quality of CDL labels, we divided the grids into train, validation and test set. The crop distribution is not similar throughout the California Central Valley crop belt. For example, Rice is grown mainly in the northern part of the California Central Valley crop belt and is rarely found in the southern parts. To make sure we create a training set that is balanced amongst classes and is also spread uniformly across space, we adopt a gridwise count based data splitting strategy. For each class, we sort the grids based on the number of pixels present of that particular class in the grid. We take intersection of the common grids amongst all classes such that the total number of grids comes out to around 60\% of the total girds we have. Following this first step we were able to filter 210 grids for training (approximately 57\%). Now after removing these training girds we again sort grid intersections classwise to create validation set such that around 20\% more is used, then the remaining are used for testing. With this approach we created a training set of 210 grids, validation set of 84 grids and test set of 73 grids. The color coded final distribution of the sets can be seen in the Fig. \ref{fig:distri2}.

\begin{figure*}[t]
    \centering
    \includegraphics[width = 0.9\textwidth]{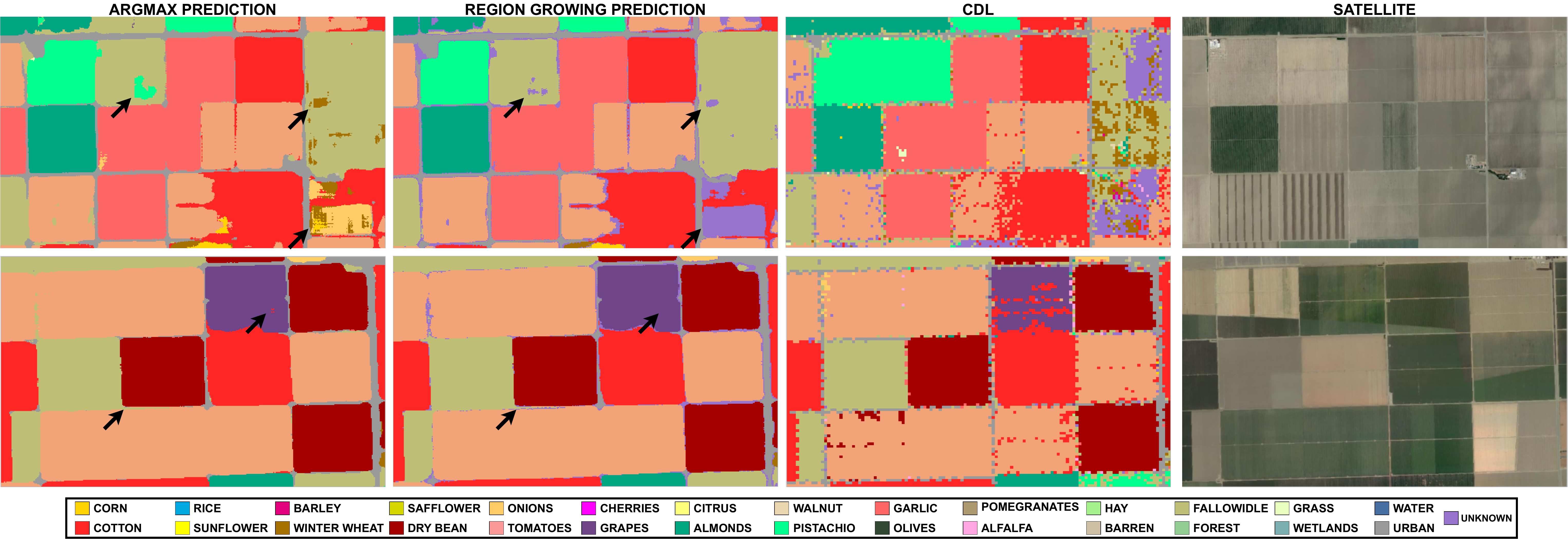}
    \caption{Argmax predictions vs Region Growing predictions on certain patches in test area which demonstrate the advantage of Region Growing over Argmax. For example in the first row, the region growing strategy has removed many speckles of winter wheat (brown) misclassified by CDL. Moreover, majority of the pixels converted to ``unknown'' by region growing lie in the boundary where the pixels due to being mixed, are more likely to be misclassified. (Arrows represent places of improvement by Region growing over Argmax)}
    \label{fig:argmaxvsrapt}
\end{figure*}

Following the approach as outlined in \cite{ghosh2021attentionaugmented}, STATT extracts patches of size 32$\times$32 pixels from the training grids. Using this input patch of size 32$\times$32, we output labels for a patch of size 16$\times$16. For this task, we use three convolutional blocks in our encoder each having two convolutional layers. Thus there are six convolutional layers having {64,64,128,128,256,256} channels and filters of size 3$\times$3. To downsample the output of the convolutional blocks STATT uses max-pooling of size 2$\times$2 after the first and second convolutional blocks. In the decoder, STATT has two convolutional blocks each of which consistsof two convolutional layers. The four convolutional layers of the decoder have {128,128,64,64} channels respectively. To upsample the output we add transposed-convolutional layers before the first and second convolutional block of the decoder having {128,64} channels respectively and kernel size of 2$\times$2. Finally, STATT has a fully-connected layer with input dimension of 64 and output dimension equal to the number of classes i.e. 28. The model was trained using the training dataset for 50 epochs and the validation performance was used as the model selection criteria.

\begin{figure}[t]
    \centering
    \includegraphics[width = 0.6\columnwidth]{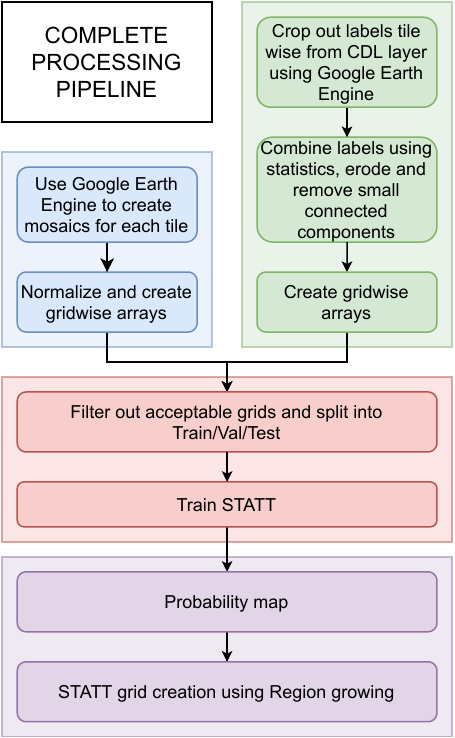}
    \caption{Flowchart of the entire pipeline used in our study.}
    \label{fig:flowchart}
\end{figure}

The output of the model are softmax probabilities over the classes for each pixel thus having shape of (16,16,34). By combining all the patches within a grid, we create probability grids of shape (16,16,34). Usually for multi-class classification the decision is made by predicting the class for which the model gives the highest probability. We refer to it as the $argmax$ prediction. In a multi-class classification setting, confusion between classes can easily occur when dealing with a large number of classes. Furthermore, class confusion also happens at the geographical boundary of different classes (e.g. fields with different crops or roads around field).

We use a region growing strategy to post-process the pixel-wise probability outputs instead of directly taking $argmax$ outputs. Specifically, for each class, the pixels that have highest probability value greater than 0.9 are considered as confident anchor pixels. Starting from these anchor pixels, we include all the pixels in their neighborhood which have at least 0.3 probability of belonging to the same class as the anchor pixels. Since the region growing strategy produces class-wise prediction maps, clashes between two or more class at certain pixels are bound to happen, in which case we assign unknown values to those pixels. We observe that majority of such clashes occur near the boundaries which is expected due to the reasons that were described above. As illustrated in Fig~\ref{fig:argmaxvsrapt}, this method is very effective in removing noise within fields and also removing confusion at boundaries by replacing them with "unknown". We store the STATT labels in arrays of shape (1098,1098) and have the naming convention as TILEID\_YEAR\_ROW\_COL\_PREPROCESSED\_STATT\_LABEL.npy.

\begin{figure*}[t]
    \centering
    \includegraphics[width = \textwidth]{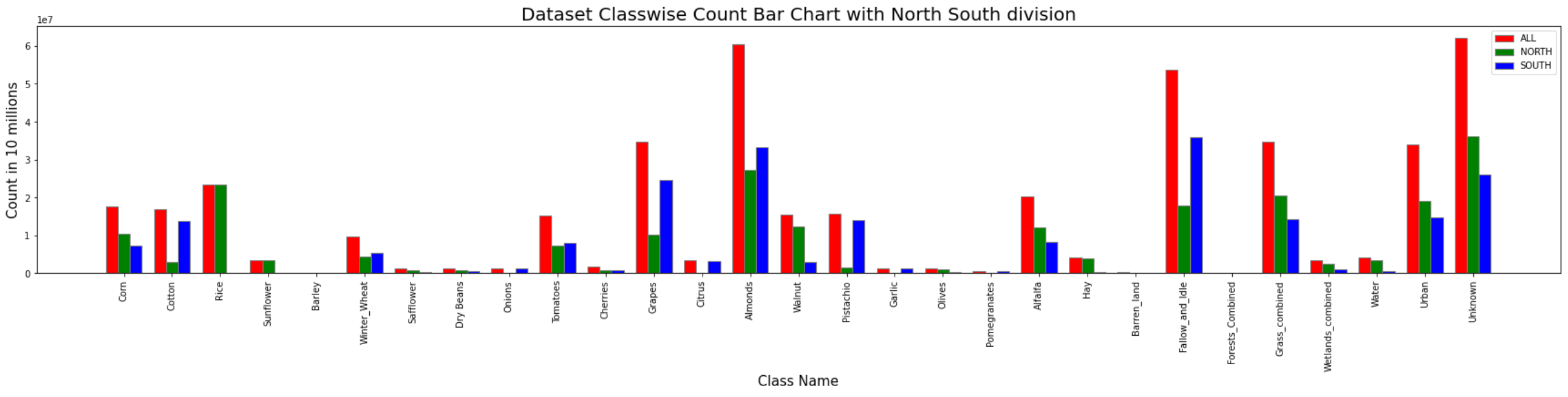}
    \caption{The classwise count of labels produced by STATT across all the acceptable grids in the form of a bar chart. The red bar represents the overall count of that class across the entire acceptable set, the green bar represents the count in the northern region and the blue bar represents the count in the southern region. The northern region is given by the tiles T10SEH, T10SEJ, T10SFG, T10SFH, T10SFJ, T10SGG and T10TEK whereas the southern region is given by the tiles T10SGF, T11SKA, T11SKV, and T11SLV. through this bar chart one can observe that there is rich crop variety in our dataset. One can also observe that some classes lie predominately in the north, such as Rice, some lie predominately in the south, such as Pistachio, and some have a good mix in both regions, such as Almonds }
    \label{fig:barchart}
\end{figure*}

\subsection{Final Dataset}

In summary, our dataset covers the entire California Central Valley Crop Belt using the 367 grids of cloud filtered multi-spectral images (each in (1098,1098,10)), and we call these image grids. For each image grid, we also provide both the raw and preprocessed CDL grid as well as STATT grid of size (1098,1098). The diagrammatic flowchart of the entire pipeline can be found in Figure \ref{fig:flowchart}. STATT labels are provided for a total of 442,456,668 pixels (\~44,000 sq. km) covering 29 classes, of which 249,946,750 pixels (\~25,000 sq. km) belong to one of the 21 crop classes and the remaining 192,509,918 pixels (\~19,000 sq. km) belong to other 8 classes including unknown. The distribution of the 29 classes in the form of a bar chart is given in Figure \ref{fig:barchart}. Figure \ref{fig:barchart} also shows the distributions class wise in the north vs south of the region, with The northern region given by the tiles T10SEH, T10SEJ, T10SFG, T10SFH, T10SFJ, T10SGG and T10TEK and the southern region given by the tiles T10SGF, T11SKA, T11SKV, and T11SLV. 

The entire dataset including Image Grids, CDL grids, preprocessed CDL grids and STATT grids for the acceptable grids as well as the Image grids and CDL grids for the rest of the entire region can be found in the link given below\footnote{\url{https://drive.google.com/drive/folders/1EnXXRHNoTyIbM-_5p-P9pH4zH3xyTqBp?usp=sharing}}.

\begin{table*}[]
\centering
\caption{Confusion matrix between STATT labels and CDL labels on test grids. X axis (column data) represents the STATT predictions and Y axis (row data) represents CDL predictions.}
\resizebox{\textwidth}{!}
{
\begin{tabular}{|l|l|cccccccccccccccccccccccccccc|}
\hline 
& & \multicolumn{28}{c|}{STATT}  \\ \hline
&Class             & \rot{90}{Corn}   & \rot{90}{Cotton}  & \rot{90}{Rice} & \rot{90}{Sunflower} & \rot{90}{Barley} & \rot{90}{Winter Wheat} & \rot{90}{Safflower} & \rot{90}{Dry Beans} & \rot{90}{Onions} & \rot{90}{Tomatoes} & \rot{90}{Cherries} & \rot{90}{Grapes} & \rot{90}{Citrus} & \rot{90}{Almonds} & \rot{90}{Walnut} & \rot{90}{Pistachio} & \rot{90}{Garlic} & \rot{90}{Olives} & \rot{90}{Pomegranates} & \rot{90}{Alfalfa} & \rot{90}{Hay} & \rot{90}{Barren} & \rot{90}{FallowIdle} & \rot{90}{ForestsCb} & \rot{90}{GrassCb} & \rot{90}{WetlandsCb} & \rot{90}{Water} & \rot{90}{Urban}    \\\hline
\multirow{28}{*}{\rot{90}{Crop Data Layer}}& \rot{0}{Corn}              & 607390 & 3178    & 4282    & 164       & 17     & 3285         & 73        & 413       & 326    & 10341    & 338      & 12767   & 3      & 8373    & 726     & 85        & 0      & 26     & 2            & 13656     & 19615  & 0           & 16406           & 0                & 2938           & 773               & 946    & 10778   \\
& \rot{0}{Cotton}            & 5629   & 3277971 & 272     & 0         & 0      & 112          & 28        & 1824      & 22     & 59832    & 9        & 10382   & 0      & 9657    & 1157    & 283       & 28     & 0      & 1            & 1777      & 92     & 1           & 6252            & 0                & 357            & 250               & 14     & 6490    \\
& \rot{0}{Rice}              & 5416   & 0       & 6879520 & 269       & 0      & 0            & 62        & 5         & 0      & 5946     & 25       & 6351    & 0      & 395     & 637     & 9         & 0      & 33     & 0            & 1126      & 891    & 0           & 13152           & 0                & 22             & 53                & 1281   & 4948    \\
& \rot{0}{Sunflower}         & 7841   & 1       & 1466    & 221533    & 0      & 36           & 2413      & 35        & 0      & 7755     & 0        & 2321    & 0      & 1813    & 4063    & 17        & 0      & 6      & 0            & 192       & 66     & 0           & 9210            & 0                & 65             & 189               & 218    & 2023    \\
& \rot{0}{Barley}            & 286    & 213     & 0       & 0         & 15238  & 31066        & 222       & 0         & 6      & 13       & 0        & 59      & 0      & 8821    & 0       & 525       & 0      & 3      & 0            & 49        & 8979   & 8           & 96979           & 0                & 8498           & 0                 & 0      & 1063    \\
& \rot{0}{Winter Wheat}      & 35205  & 356     & 9       & 0         & 15     & 1145627      & 6270      & 4886      & 23840  & 3049     & 1051     & 3189    & 0      & 4996    & 313     & 267       & 2472   & 20     & 143          & 7324      & 28273  & 0           & 569444          & 0                & 85098          & 50                & 204    & 5277    \\
& \rot{0}{Safflower}         & 40897  & 21      & 351     & 2139      & 0      & 617          & 383917    & 15        & 87569  & 48333    & 6        & 5438    & 0      & 596     & 616     & 56        & 12049  & 4      & 0            & 16077     & 525    & 0           & 64979           & 0                & 16257          & 13                & 23     & 2227    \\
& \rot{0}{Dry Beans}         & 1529   & 2009    & 38      & 7         & 0      & 13148        & 2         & 94384     & 90     & 572      & 71       & 1350    & 0      & 400     & 33      & 1         & 888    & 0      & 0            & 2744      & 3380   & 0           & 4315            & 0                & 55             & 81                & 1      & 972     \\
& \rot{0}{Onions}            & 9514   & 50      & 0       & 0         & 2      & 1001         & 8001      & 0         & 201520 & 16235    & 0        & 305     & 0      & 15      & 0       & 1         & 3979   & 0      & 0            & 555       & 16     & 0           & 18561           & 0                & 24             & 0                 & 0      & 945     \\
& \rot{0}{Tomatoes}          & 36578  & 30306   & 446     & 639       & 0      & 655          & 1598      & 345       & 19816  & 2421624  & 76       & 5568    & 0      & 8861    & 172     & 166       & 323    & 0      & 2            & 4824      & 384    & 0           & 42105           & 0                & 600            & 60                & 8      & 5831    \\
& \rot{0}{Cherries}          & 12     & 155     & 0       & 0         & 0      & 6            & 0         & 0         & 0      & 0        & 45327    & 22724   & 26     & 5012    & 4857    & 2124      & 0      & 0      & 324          & 60        & 0      & 0           & 15426           & 0                & 5259           & 0                 & 0      & 769     \\
& \rot{0}{Grapes}            & 14290  & 12029   & 0       & 0         & 0      & 1184         & 31        & 23        & 1956   & 18507    & 1977     & 2555800 & 14     & 126264  & 4362    & 14492     & 0      & 2255   & 6647         & 8998      & 851    & 0           & 64457           & 0                & 11970          & 1                 & 11     & 8969    \\
& \rot{0}{Citrus}            & 0      & 0       & 0       & 0         & 0      & 0            & 0         & 0         & 0      & 1967     & 0        & 739     & 145307 & 620     & 434     & 9         & 3      & 4063   & 9            & 371       & 0      & 0           & 17482           & 0                & 931            & 0                 & 0      & 1426    \\
& \rot{0}{Almonds}           & 9401   & 5609    & 25      & 231       & 0      & 16234        & 1287      & 3807      & 14     & 2206     & 3129     & 113893  & 1560   & 7592014 & 85562   & 29228     & 2715   & 6785   & 3114         & 34172     & 427    & 3           & 285501          & 0                & 99096          & 138               & 31     & 40524   \\
& \rot{0}{Walnut}            & 6270   & 2281    & 2811    & 3124      & 0      & 508          & 4707      & 110       & 0      & 72       & 3787     & 269005  & 0      & 91716   & 2792583 & 43332     & 0      & 399    & 694          & 44839     & 1568   & 4           & 141135          & 0                & 13121          & 5735              & 686    & 40591   \\
& \rot{0}{Pistachio}         & 332    & 290     & 0       & 0         & 0      & 73           & 1         & 17        & 8      & 7        & 2365     & 14843   & 14     & 48152   & 4853    & 2273489   & 1      & 210    & 8372         & 2236      & 0      & 0           & 235936          & 0                & 63432          & 0                 & 11     & 11139   \\
& \rot{0}{Garlic}            & 1063   & 23      & 0       & 0         & 0      & 8566         & 91        & 464       & 6964   & 604      & 11       & 65      & 6      & 48      & 0       & 8         & 302008 & 0      & 0            & 131       & 0      & 0           & 11529           & 0                & 102            & 43                & 0      & 913     \\
& \rot{0}{Olives}            & 429    & 0       & 3       & 0         & 0      & 0            & 0         & 0         & 0      & 0        & 0        & 7524    & 10259  & 11251   & 1201    & 923       & 0      & 149718 & 0            & 516       & 12     & 6           & 24853           & 0                & 3977           & 0                 & 0      & 4003    \\
& \rot{0}{Pomegranates}      & 2      & 4611    & 0       & 0         & 0      & 0            & 0         & 2         & 1      & 2130     & 3140     & 12175   & 7      & 30375   & 56      & 17403     & 5      & 123    & 94072        & 361       & 0      & 0           & 50850           & 0                & 705            & 87                & 54     & 2882    \\
& \rot{0}{Alfalfa}         & 31847  & 1599    & 322     & 23        & 0      & 11704        & 4852      & 7         & 6590   & 6543     & 860      & 27857   & 191    & 31837   & 5306    & 1852      & 1237   & 1087   & 130          & 1372785   & 14676  & 0           & 31217           & 0                & 8036           & 2733              & 271    & 12958   \\
& \rot{0}{Hay}               & 13640  & 158     & 417     & 1407      & 0      & 26405        & 217       & 404       & 92     & 501      & 641      & 39422   & 624    & 25357   & 6998    & 1070      & 0      & 4805   & 6            & 144939    & 152060 & 1           & 96765           & 0                & 107317         & 171               & 120    & 20725   \\
& \rot{0}{Barren}       & 0      & 150     & 51      & 0         & 12     & 0            & 0         & 0         & 0      & 14       & 1        & 278     & 0      & 359     & 144     & 75        & 0      & 40     & 1            & 42        & 4      & 20040       & 97624           & 0                & 106082         & 633               & 8816   & 38194   \\
& \rot{0}{FallowIdle}   & 11712  & 17863   & 164556  & 986       & 44     & 34016        & 13008     & 7834      & 2990   & 71154    & 805      & 81176   & 76     & 125921  & 84253   & 89519     & 1102   & 7941   & 1363         & 32032     & 17392  & 1351        & 6817036         & 0                & 573836         & 7230              & 21056  & 262682  \\
& \rot{0}{ForestsCb}  & 6761   & 0       & 4135    & 0         & 0      & 3            & 153       & 0         & 0      & 0        & 1721     & 7832    & 10142  & 2422    & 451     & 0         & 0      & 115216 & 0            & 39529     & 15     & 2           & 20653           & 0                & 257            & 731               & 48     & 1281    \\
& \rot{0}{GrassCb}    & 12258  & 348     & 1042    & 0         & 823    & 11029        & 620       & 6         & 0      & 113      & 3004     & 81509   & 640    & 92248   & 17723   & 19635     & 0      & 26063  & 179          & 92681     & 141563 & 409         & 1307273         & 0                & 4677219        & 11878             & 2201   & 79275   \\
& \rot{0}{WetlandsCb} & 29639  & 614     & 32684   & 0         & 0      & 21           & 3         & 0         & 0      & 48       & 478      & 24263   & 0      & 811     & 7426    & 1226      & 0      & 31     & 2            & 8927      & 187    & 44          & 83276           & 0                & 76186          & 103289            & 19880  & 7644    \\
& \rot{0}{Water}             & 271    & 0       & 1658    & 0         & 0      & 8            & 0         & 0         & 0      & 0        & 0        & 130     & 0      & 9       & 51      & 0         & 0      & 6      & 0            & 215       & 3      & 958         & 4940            & 0                & 2620           & 1748              & 920775 & 2100    \\
 & \rot{0}{Urban}             & 2469   & 9142    & 21946   & 574       & 0      & 3466         & 675       & 144       & 621    & 7711     & 384      & 28072   & 907    & 48990   & 22374   & 12213     & 971    & 4083   & 172          & 18503     & 2053   & 3045        & 201794          & 0                & 177117         & 2542              & 13400  & 2747186\\\hline
\multicolumn{30}{|c|}{FallowIdle - Fallow and Idle land; ForestsCb - Forests Combined; GrassCb - Grass Combined; WetlandsCb - Wetlands Combined}\\\hline 
\end{tabular}}
\label{Tab:confusionmatrix}
\end{table*}

\begin{table*}[h!]
\footnotesize
\centering
\caption{Precision, Recall, and F1-Score of STATT labels in the test region with CDL as groundtruth. We also mention support(in pixels), that is the count classwise of CDL labels used during evaluation (10000 pixels equals 1 sq.km). }
\begin{tabular}{|l|cccc|l|cccc|}
\hline
\multicolumn{10}{|c|}{STATT}                                                                                                                  \\ \hline 
CLASS           & Precision & Recall & F1-score & Support  & CLASS             & Precision & Recall                   & F1-score & Support    \\ \hline
Fallow and Idle & 0.6587    & 0.8069 & 0.7253   & 8448934  & Safflower         & 0.8965    & 0.5623                   & 0.6911   & 682725     \\
Almonds         & 0.9172    & 0.9107 & 0.9139   & 8336706  & Hay               & 0.3869    & 0.2360                    & 0.2932   & 644262     \\
Rice            & 0.9668    & 0.9941 & 0.9803   & 6920141  & Wetlands Combined & 0.7462    & 0.2604                   & 0.386    & 396679     \\
Grass Combined  & 0.7742    & 0.7109 & 0.7412   & 6579739  & Garlic            & 0.9214    & 0.9079                   & 0.9146   & 332639     \\
Walnut          & 0.9167    & 0.8050 & 0.8572   & 3469078  & Barren            & 0.7746    & 0.0735                   & 0.1343   & 272560     \\
Cotton          & 0.9730    & 0.9691 & 0.9710   & 3382440  & Sunflower         & 0.9586    & 0.8479                   & 0.8999   & 261263     \\
Urban           & 0.8265    & 0.8248 & 0.8257   & 3330554  & Onions            & 0.5718    & 0.7729                   & 0.6573   & 260724     \\
Grapes          & 0.7663    & 0.8952 & 0.8258   & 2855088  & Pomegranates      & 0.8164    & 0.4295                   & 0.5628   & 219041     \\
Pistachio       & 0.9065    & 0.8528 & 0.8788   & 2665781  & Olives            & 0.4636    & 0.6974                   & 0.557    & 214675     \\
Tomatoes        & 0.9018    & 0.9383 & 0.9197   & 2580987  & Forests Combined  & 0.0000    & 0.0000                   & 0.0000   & 211352     \\
Winter Wheat    & 0.8753    & 0.5944 & 0.7080   & 1927378  & Citrus            & 0.8559    & 0.8382                   & 0.8469   & 173361     \\
Alfalfa         & 0.7422    & 0.8708 & 0.8013   & 1576520  & Barley            & 0.9435    & 0.0886                   & 0.162    & 172028     \\
Water           & 0.9300    & 0.9843 & 0.9564   & 935492   & Dry Beans         & 0.8227    & 0.7487                   & 0.7839   & 126070     \\
Corn            & 0.6819    & 0.8472 & 0.7557   & 716901   & Cherries          & 0.6550    & 0.4440                    & 0.5293   & 102081     \\ \hline \hline

OVERALL         & Precision & Recall & F1-score & Support  & CROPS ONLY        & Precision & Recall                   & F1-score & Support    \\ \hline
MEAN            & 0.7732    & 0.6754 & 0.6885   & 57795199 & MEAN:             & 0.8067    & 0.7262                   & 0.7386   & 37,619,889 \\
Weighted MEAN   & 0.834     & 0.8307 & 0.8251   & 57795199 & Weighted MEAN     & 0.8882    & 0.8699                   & 0.8731   & 37,619,889 \\
ACCURACY:       &           &        & 0.8307   &          & ACCURACY          &           &                          & 0.9303   &            \\ \hline           
\end{tabular}
\label{Tab:scores}
\end{table*}

\section{Evaluation}
\label{Sec:Eval}
In this section we present the quantitative analysis of the results of our approach on the test regions. Using the STATT and CDL labels we get the confusion matrix as shown in Table~\ref{Tab:confusionmatrix} for the pixels where both STATT and CDL are not unknown. We observe that out of the total 57,795,199 pixels, SAATT and CDL labels differ in 9,785,767 pixels (16.93\%). Focusing only on pixels that are labeled as crop, disagreement drops to 6.97\%. Table~\ref{Tab:scores} shows precision, recall, and F1-scores for all classes while treating pre-processed CDL labels as ground truth. We notice that F1-score is usually high for classes that have high support (see left half of Table 2) and usually low for classes that have low support (right half of Table 2). As we discuss in the following, STATT labels are generally more accurate than those provided by CDL.

\begin{figure*}[t]
    \centering
    \includegraphics[width = 0.9 \textwidth]{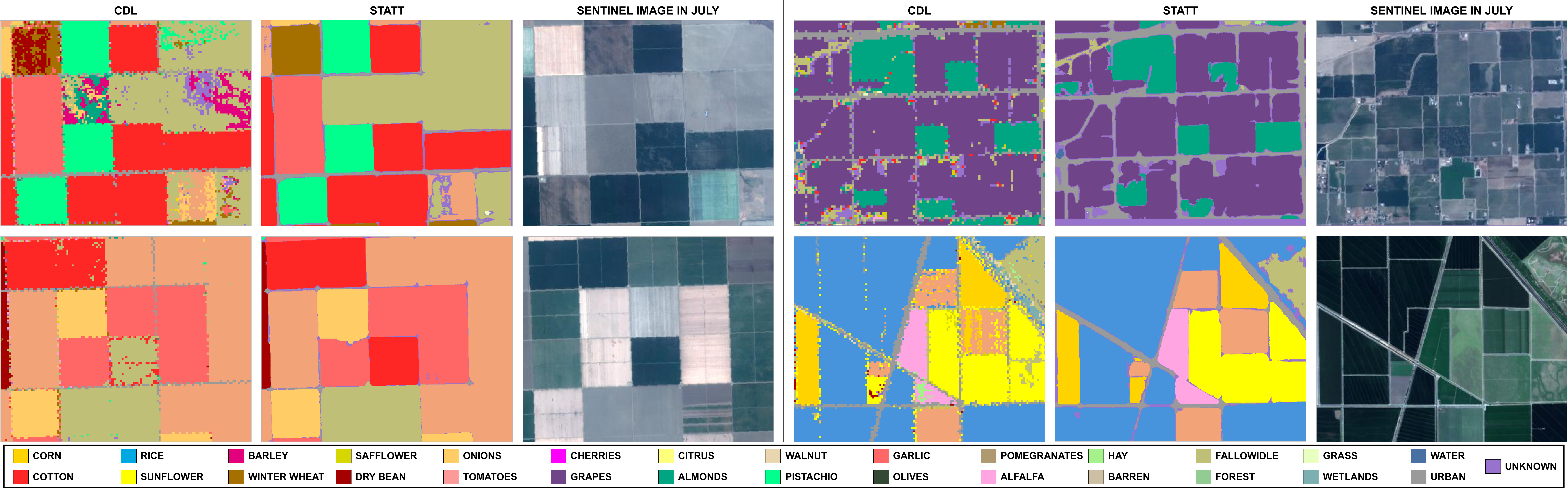}
    \caption{Segmentation map comparisons on some patches from CDL and STATT in the test regions, Each triplet shown depicts a situation where STATT produces better labels. For description on each triplet please refer to Section \ref{Sec:Eval}}
    \label{fig:segmentationmaps}
\end{figure*}

Fig.~\ref{fig:segmentationmaps} shows a comparison of the segmentation maps of STATT and the corresponding patch from the CDL layer. In all four triplets, We notice that STATT generally performs much better in detecting boundaries and removing noise. In the first triplet of the first column, we can see how a noisy field is replaced with a smooth prediction of fallow land. In the second we can see how a fallow prediction by CDL is replaced with cotton by STATT and it can be verified in the third image of the triplet, an image in July, that the field cannot be fallow as there is a crop present. In the first triplet of the second column, one can observe removal of erroneous cotton speckles present in the CDL map, and in the final triplet we can see smoothing of multiple fields by the STATT map over the region.

Further, we analyze the pixels where our map does not match with CDL. We have noticed errors in the CDL layer at numerous locations throughout the crop belt which are mainly of two types:
\begin{itemize}
    \item incorrect labeling of complete (or large parts of) fields
    \item spatially discontinuous label prediction (i.e., label of a pixel differs from its surrounding pixels that all belong to a different class)
\end{itemize}

\begin{figure*}[t]
    \centering
    \includegraphics[width =0.9 \textwidth]{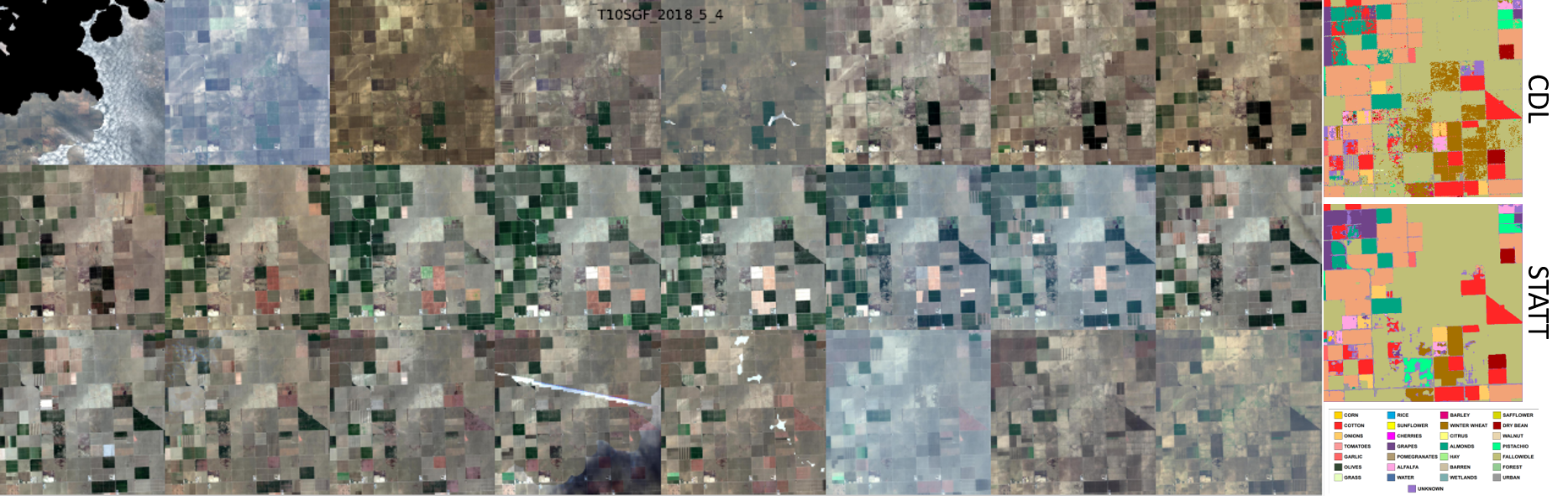}
    \caption{Visual Analysis of Grid T10SGF\_2018\_5\_4. One can observe from the visual images that many fields in this region are fallow throughout the year but CDL labels them as Winter Wheat (such as the field next to the triangular Cotton (red) field). However STATT does not make these mistakes and correctly labels the region as Fallow land.}
    \label{fig:fallowvswinterwheat}
\end{figure*}

In the next few paragraphs we systemically discuss and analyze these cases on some of the fields within California Crop belt. First, we visually analyze and show the results of a sample of patches by looking at the satellite images over the entire year. Next, we provide visual analysis of a sample of NDVI time series. We assume that the NDVI time series of pixels where STATT and CDL agree are correct and we show that on pixels where STATT disagrees with CDL, the NDVI time series are similar to those pixels that have the same label as provided by STATT and where STATT and CDL agree. Finally, we provide a comprehensive (qualitative) analysis of all pixels where STATT and CDL disagree by quantifying the number of pixels that have NDVI series closer to the agreement NDVI series for each set of labels.

\subsection{Visual analysis of a sample of patches}
The first way to compare our dataset and CDL is  to visually inspect images over time, check the growing time, rate of greenness and harvest time to assign a label to the field, and then check whether CDL or STATT is correct. Although this method cannot be scaled to every field due to substantial manual effort and expertise needed regarding crop growing patterns,  we can still use this approach to verify disagreement between CDL and STATT where one predicts fallow and idle land and the other predicts a crop. Since no crop is grown year around in a fallow or idle land, it should be easy to verify the correct prediction. We observed numerous cases in the California Central Valley crop belt where CDL predicted a crop and STATT predicted fallow land. On further investigation as to why so much confusion has happened, we observed in the CDL that some classes, such as almonds, grapes, and barley, are just carried over from the year before without new analysis. This is mainly a problem when the farmer decides not to plant a crop in that area that year or a drought occurs in that area that year and so in reality the pixel is fallow land but CDL will label it as a crop. In such cases even though STATT may label the pixel as Fallow land it will be taken as an error with respect to CDL. In particular, 2018 was a dry year for the California Central Valley and many farmers were not able to irrigate fields in 2018 that were wet and productive in 2017. We found this to be the main reason as to why the F1 score for Fallow and Idle land is low in the table in spite of a high count. An example of what was just described can be seen in Fig.~\ref{fig:fallowvswinterwheat}. In this grid, We can observe numerous fields as fallow throughout the year (such as the one one next to the triangle shaped cotton field) but CDL labels them as crop (the field next to the triangle cotton field is labelled as winter wheat by CDL). Through this first method of visual analysis, we were able to verify many cases of fallow vs crop disagreement, in which the STATT prediction of fallow seems appropriate.

\begin{figure*}[t]
    \centering
    \includegraphics[width = .95\textwidth]{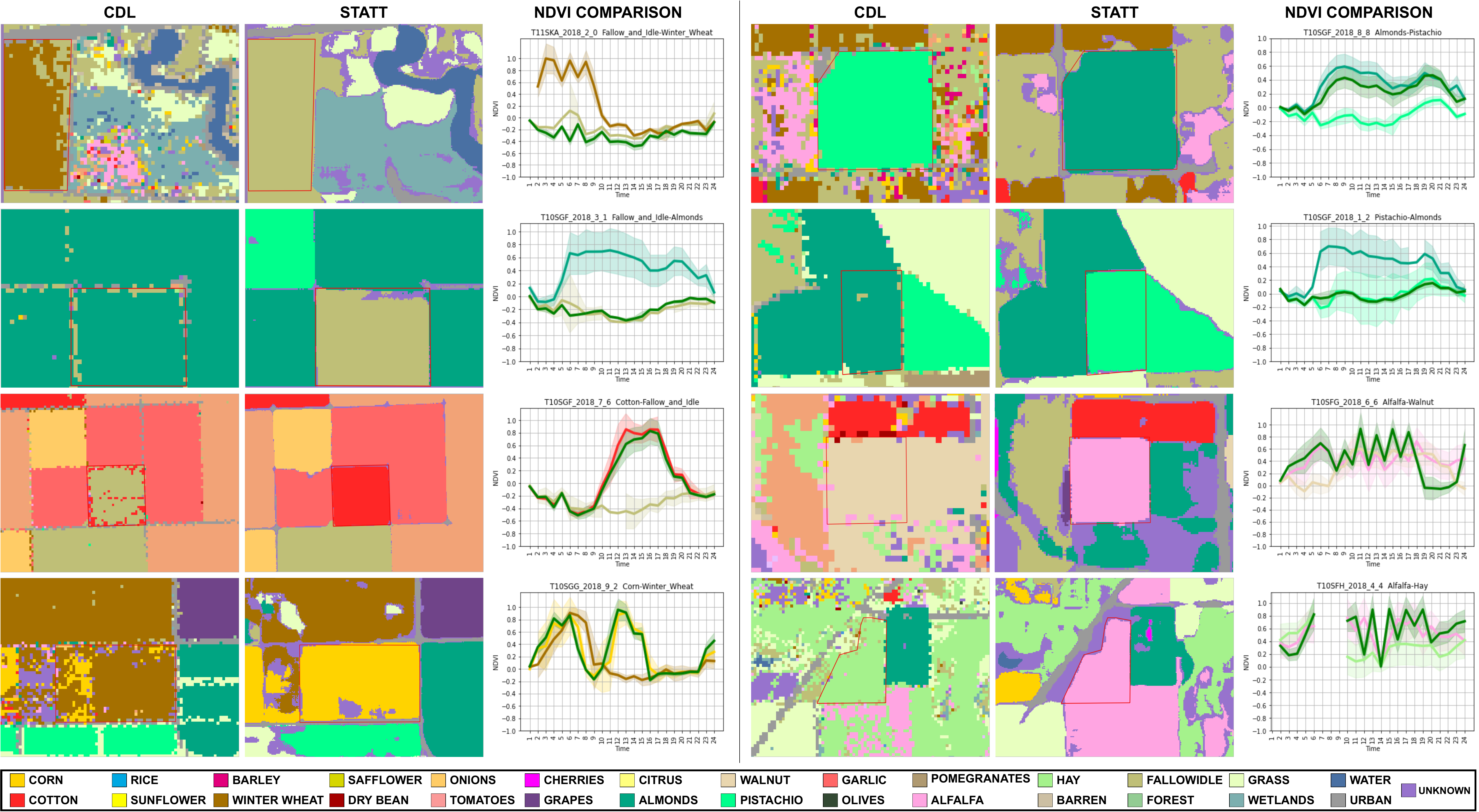}
    \caption{Comparison of NDVI for some fields with disagreeemnt between CDL labels and STATT labels. Each triplet denotes a case where the labels produced by STATT are better than CDL, as the series of the field (represented in Green) lies closer to the series of the class denoted by STATT.}
    \label{fig:fieldanalysis}
\end{figure*}

\subsection{Visual analysis of a sample of NDVI time series} 
Here we resolve the disagreement between STATT and CDL labels by analyzing the NDVI time series of the field in question (i.e. a continuous region where STATT and CDL differs. In each disagreement there are two classes in analysis, the CDL prediction class and the STATT prediction class. If the NDVI series of field lies closer to the characteristic NDVI series of the STATT prediction class than the CDL prediction class then we can say that the field is actually the STATT prediction class and vice versa. Now the question arises, how do we obtain this characteristic NDVI series of different classes in our dataset? To get the characteristic NDVI series we take the median (timestamp wise) of the NDVI series for pixels agreeing with the class of interest in the grid where the field of interest is located. What we mean by pixels of agreement are those pixels where CDL and STATT agree, i.e predict the same class at that pixel. we use only agreement pixels within the grid of the field because we found that across grids crops have different NDVI series due to local farmer practices, weather conditions and cloud cover patterns. We also use median to get the characteristic NDVI series due to the fact that it is better at handling outliers when compared to other strategies such as mean or mode. 
    
We now plot the characteristic NDVI series for both the CDL prediction class and the STATT prediction class on the same graph. We then plot the median NDVI series timestamp wise of all the pixels in the field of interest on the same graph and check which characteristic NDVI series it lies closer to. If the NDVI series of the field is closer to the characteristic series of the class labeled by STATT compared to the characteristics series of the class labeled by CDL, then we can say that STATT label was correct, and vice versa. We found that in a vast majority of cases, whenever there is a field of disagreement, the NDVI series of the field lies closer to the STATT prediction class signature than the CDL prediction class signature. Fig.~\ref{fig:fieldanalysis} shows 8 triplets for some fields where we conducted this method of analysis. The first image in the triplet is the CDL prediction and the second image is the STATT prediction, and in each of these images there is a red boundary denoting the field of interest. One can observe that in all the triplets, the predicted class within the field of interest (i.e the red boundary) differs between the CDL image and the STATT image. The third image is the NDVI plot of the three timeseries described before, i.e the CDL prediction class characteristic NDVI series (denoted by the plot in the color of the CDL prediction class), the STATT prediction class characteristic NDVI series (denoted by the plot in the color of the STATT prediction class) and the NDVI series of the field of interest (denoted by the signature in the green color).

To give a better understanding let us look at the top leftmost triplet of Fig.~\ref{fig:fieldanalysis} in detail. This triplet shows a field which has been labelled as winter wheat by CDL but as fallow land by STATT. In the third image we can see the NDVI series signature of winter wheat in dark brown and that of fallow land in light brown (note that the colors of the lines are the same colors of the fields in the images). We can also see the green line plot in the image which denotes that of the field's NDVI series. It can clearly be seen that the green plot lies much closer to the light brown plot (fallow land) and so we can say the STATT prediction of this field as fallow land is correct. Similarly the top rightmost triplet shows a field with pistachio (CDL prediction) vs Almond (STATT predication) and from the plot we can see that the green line lies closer to the greenish blue line (Almond) than when compared to the light blue line (Pistachio) showing that the STATT prediction in this field is correct. All other examples in Figure 9 similarly show that the green line lies closer to the STATT class NDVI line. The top two triplets of the right column show cases where CDL predicts a crop class and STATT predicts fallow land. The bottom next triplet of the right column show cases where CDL predicts fallow land but STATT predicts a non fallow class (Cotton). The last triplet in this column shows winter wheat vs corn. The top two triplets of the left column showcase confusion of almonds and pistachio in both ways between the two tree crops which are of high importance in the California region. The next triplet shows walnut vs alfalfa and the final triplet shows alfalfa vs hay. 

\begin{figure}[]
    \centering
    \includegraphics[width = \columnwidth]{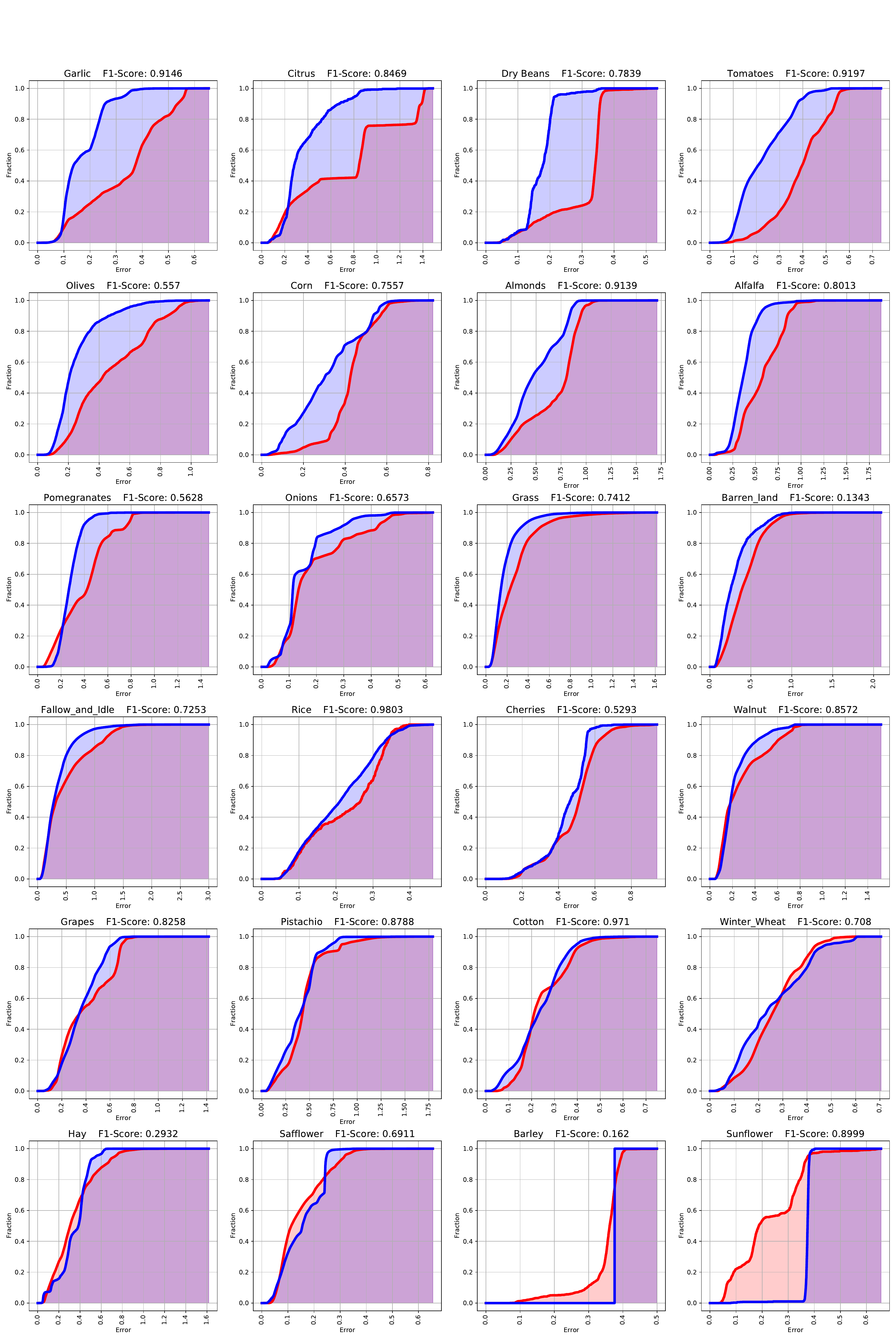}
    \caption{
    Class wise Area under the curve plots. The x axis represents the NMSE and the y axis represents the $Score(\cdot)$ value for the corresponding NMSE from x axis. 
    The red and blue curves represent CDL and STATT, respectively. 
    In each plot we also mention the F1-Score of that class 
    as a measure of agreement between CDL and STATT. We can see that in classes with high F1 such as Garlic, where agreement is high, and in those with low F1 such as Olives, where agreement is low, STATT has a better curve than CDL. 
    }
    \label{fig:auc}
\end{figure}

\begin{table}[]
    \footnotesize
        \caption{Area under the curves (in Fig.~\ref{fig:auc}) 
        for each class along with number of agreement pixels and F1 score of class with CDL as groundtruth. }
    \begin{tabular}{|c|ccc|c|c|}
    \hline
                         & \multicolumn{3}{c|}{AREA UNDER THE CURVE}      & Count  &   \\ \hline
Class                & CDL             & STATT           & Difference & Agreement & F1-Score \\ \hline
Garlic               & 0.4899          & \textbf{0.7409} & 0.2510     & 173908    &  0.9146 \\
Citrus               & 0.5097          & \textbf{0.7555} & 0.2458     & 26018     &  0.8469 \\
Dry Beans            & 0.4388          & \textbf{0.6803} & 0.2415     & 59647     &  0.7839 \\
Tomatoes             & 0.4708          & \textbf{0.6927} & 0.2218     & 1344877   &  0.9197 \\
Olives               & 0.5745          & \textbf{0.7741} & 0.1995     & 52639     &  0.5570 \\
Corn                 & 0.4850          & \textbf{0.6177} & 0.1327     & 68350     &  0.7557 \\
Almonds              & 0.5959          & \textbf{0.7072} & 0.1112     & 2867928   &  0.9139 \\
Alfalfa              & 0.6933          & \textbf{0.8000} & 0.1067     & 477421    &  0.8013 \\
Pomegranates         & 0.7310          & \textbf{0.8114} & 0.0804     & 71381     &  0.5628 \\
Onions               & 0.7007          & \textbf{0.7628} & 0.0621     & 146608    &  0.6573 \\
Grass Combined               & 0.8334          & \textbf{0.8910} & 0.0576     & 1336303   &  0.7412 \\
Barren         & 0.7997          & \textbf{0.8550} & 0.0552     & 8902      &  0.1343 \\
Fallow and Idle    & 0.8396          & \textbf{0.8825} & 0.0429     & 2451705   &  0.7253 \\
Rice                 & 0.5055          & \textbf{0.5481} & 0.0426     & 501687    &  0.9803 \\
Cherries             & 0.4864          & \textbf{0.5277} & 0.0413     & 28185     &  0.5293 \\
Walnut               & 0.8223          & \textbf{0.8519} & 0.0296     & 369057    &  0.8572 \\
Grapes               & 0.7197          & \textbf{0.7465} & 0.0269     & 435258    &  0.8258 \\
Pistachio            & 0.7559          & \textbf{0.7816} & 0.0257     & 1106181   &  0.8788 \\
Cotton               & 0.6727          & \textbf{0.6918} & 0.0192     & 2481209   &  0.9710 \\
Winter Wheat        & 0.6292          & \textbf{0.6403} & 0.0111     & 372512    &  0.7080 \\
Hay                  & \textbf{0.7887} & 0.7879          & -0.0009    & 28640     &  0.2932 \\
Safflower            & \textbf{0.7749} & 0.7638          & -0.0112    & 125555    &  0.6911 \\
Barley               & \textbf{0.3051} & 0.2474          & -0.0577    & 424       &  0.1620 \\
Sunflower            & \textbf{0.6578} & 0.4333          & -0.2246    & 3842      &  0.8999 \\  \hline
MEAN                 & 0.6440          & \textbf{0.7195} &            &          & \\ \hline
    \end{tabular}
    \label{Tab:auc}
\end{table}

\subsection{Comprehensive analysis of all pixels where STATT and CDL disagree}

The previous two methods prove to be very useful while doing field analysis and using a combination of the two we can show for each field who is correct. However, neither of these methods give a global perspective nor do they quantify how much better STATT's predictions are when compared to those of CDL's predictions. To address this issue, we use a third method of analysis in which we devise a function to measure closeness of pixels to ground reality, and after plotting the function, use area under the curve to establish which strategy, i.e. CDL or STATT, is better.
    
After obtaining the map for STATT and CDL, we calculate the characteristic NDVI series gridwise for each class using the agreement pixels as described in the previous section. Now we consider a characteristic series to be valid only if there are at least $T$ pixels in agreement in the grid, and $T$ is set to 100 in this work. 
Now for each pixel of disagreement for each class we calculate the Normalised Mean Square Error (NMSE) with the characteristic NDVI series and the NDVI series of the the pixel of disagreement. 
For each strategy (CDL or STATT), we first sort all the disagreement pixels according to their NMSE. Then we compute $Score(E)$ for each strategy, which is defined to be the proportion of disagreement pixels with NMSE less than a particular error $E$ over all the disagreement pixels, i.e., $Score()$, for a particular error (E) as follows:
\begin{equation}
    \small
    Score(E) = \frac{\#\, disagreement\, pixels\, in\, strategy\,with\, NMSE\, less\, than\, E}{Total\, No.\, of\, pixels\, of\, disagreement\, in\, strategy\,} 
    \label{eq:score}
\end{equation}

where $strategy$ represents either CDL or STATT. 

What the function Score represents is the fraction of total disagreements pixels whose NMSE lies below a set threshold error denoted by E. The notion behind this function is that, the closer the NMSEs of the disagreement pixels are to zero, the faster Score rises as E rises. At the max error, the Score will be 1, as all NMSEs lie below the threshold. Our hypothesis is that the STATT disagreement pixels have lower errors and so Score will rise faster for STATT when compared to CDL. As a result, STATT will reach a higher Score faster and will thus have more area under the plot of Score until the max error. A plot of this function is constructed for each class with E starting from 0 and ranging up til maximum NMSE error recorded for that class, which we denote as $E_{max}$. Please note that $E_{max}$ changes from class to class and that $E_{max}$ could come from either a CDL disagreement pixel or a STATT disagreement pixel. 
We then calculate Area under the curve as follows:

\begin{equation}
    Area_{strategy} = (\int_{0}^{E_{max}} Score(E)\, dE ) / E_{max}
    \label{eq:area}
\end{equation}

We divide the area of each plot by $E_{max}$ to keep it within the range (0,1).The plots of curve for each class can be seen in Fig.~\ref{fig:auc}, with Blue representing STATT and red representing CDL. The areas under these curves for each strategy  are summarized in Table \ref{Tab:auc}. We can see that STATT has a higher area when compared to CDL in almost all the classes. STATT has lower area for a small number of classes which have low agreement count. We also see from the figure that in a lot of classes the blue line lies above the red line throughout the plot. This experiment solidifies our claim that STATT labels are closer to the ground reality than when compared to the labels provided by CDL. 

\section{Conclusion}
In this paper we presented CalCROP21, a georeferenced data set for a diverse array of crops grown in the Central Valley of California.  This dataset contains multi spectral Sentinel imagery along with crop labels at 10m resolution for year 2018 that are derived using a novel spatial-temporal deep learning method that makes use of noisy  CDL labels available at 30m resolution.  Our extensive  analysis of this dataset demonstrates the superiority of our dataset over  CDL. We have also released our processing pipeline and associated datasets that can be used by the community to generate crop labels for other years and for creating similar data sets for other parts of US.  We anticipate this dataset will catalyze the innovation in machine learning research  on remote sensing data (e.g., classifying multiple imbalanced classes and modeling heterogeneous data over space), and also enable the use of this information for studying crop distribution and its implications by the agricultural community.

\section{Acknowledgement}
This work was funded by the NSF awards 1838159 and 1739191. Rahul Ghosh is supported by UMII MNDrive Graduate Fellowship. Access to computing facilities was provided by the Minnesota Supercomputing Institute.

\bibliographystyle{ACM-Reference-Format}
\bibliography{main}

\begin{table*}[t!]
\begin{tabular}{|c|cc|}
\hline
NMSE SCATTERPLOT STATISTICS & ALL DISAGREEMENT & CROP DISAGREEMENT \\ \hline
TOTAL POINTS & 2681881 & 2027776\\
STATT BETTER & 1672366 & 1323883\\
CDL BETTER & 1009515 & 703893\\
STATT BETTER MEAN & 0.1321 & 0.1567\\
CDL BETTER MEAN & 0.0857 & 0.1091\\ \hline
\end{tabular}
\caption{NMSE scatterplots}
\label{Tab:nmsescatterplot}
\end{table*}

\newpage

\section{Appendix}
\label{Sec:Appendix}

\subsection{\textbf{NMSE analysis}} 
In this section we propose another method of analysis, estimated F1 score analysis method. In this method we try to estimate the actual F1 score of CDL and STATT based on proximity of NDVI series of pixels with respect to the characteristic NDVI series using Normalised mean square error.
    
Like the AUC analysis we calculate the characteristic NDVI series gridwise for each class using the agreement pixels and consider a characteristic series to be valid only if there are at least 'T' pixels in agreement in the grid. Once again in this experiment, we set 'T' to be 100. Once again for each pixel of disagreement for each class we calculate the mean square error with the characteristic NDVI series and the NDVI series of the the pixel of disagreement and store all these scores and then normalise each after all disagreement pixels series errors have been calculated. We then set a threshold below which the NMSE should be for a pixel so that it can be considered for F1 score calculation. We set this threshold by finding that threshold at which at least 25 to 45 percent of the disagreement pixels have an NMSE score less than that threshold. Note that in this step we are not distinguishing between CDL and STATT pixels, but consider overall all the pixels of disagreement. Now after setting the threshold, we then find out how many of these pixels that cross the threshold are from CDL and how many are from STATT. We also have the number of agreement pixels in that grid as well and also number of disagreement pixels from STATT and CDL for the class and the grid in question. Using these numbers we can calculate the F1 score for CDL and STATT separately using the formula:
    
\begin{equation}
    F1\_Score_{strategy} = \frac{2 \times Precision_{strategy} \times Recall_{strategy}}{Precision_{strategy} + Recall_{strategy}} 
\end{equation}

\begin{equation} 
    Precision_{strategy} = \frac{count\_thresholded_{strategy} + count\_agreement}{count\_disagree_{strategy} + count\_agreement}
\end{equation}
\begin{equation} 
    Recall_{strategy} = \frac{count\_thresholded_{strategy} + count\_agreement}{count\_thresholded_{total} + count\_agreement}
\end{equation}

where $strategy$ represents either CDL or STATT. 
We conducted this analysis on all classes for which USDA reports statistics and the results can be found in Table. \ref{Tab:nmsetableResults}. As can be seen from the table STATT has a higher F1 score when compared to CDL on all but 3 classes. The table also verifies our fallow land proposition by a significantly higher f1 score.

\subsection{NMSE Scatterplot}

\begin{figure*}[t!]
    \centering
    \includegraphics[width = \textwidth]{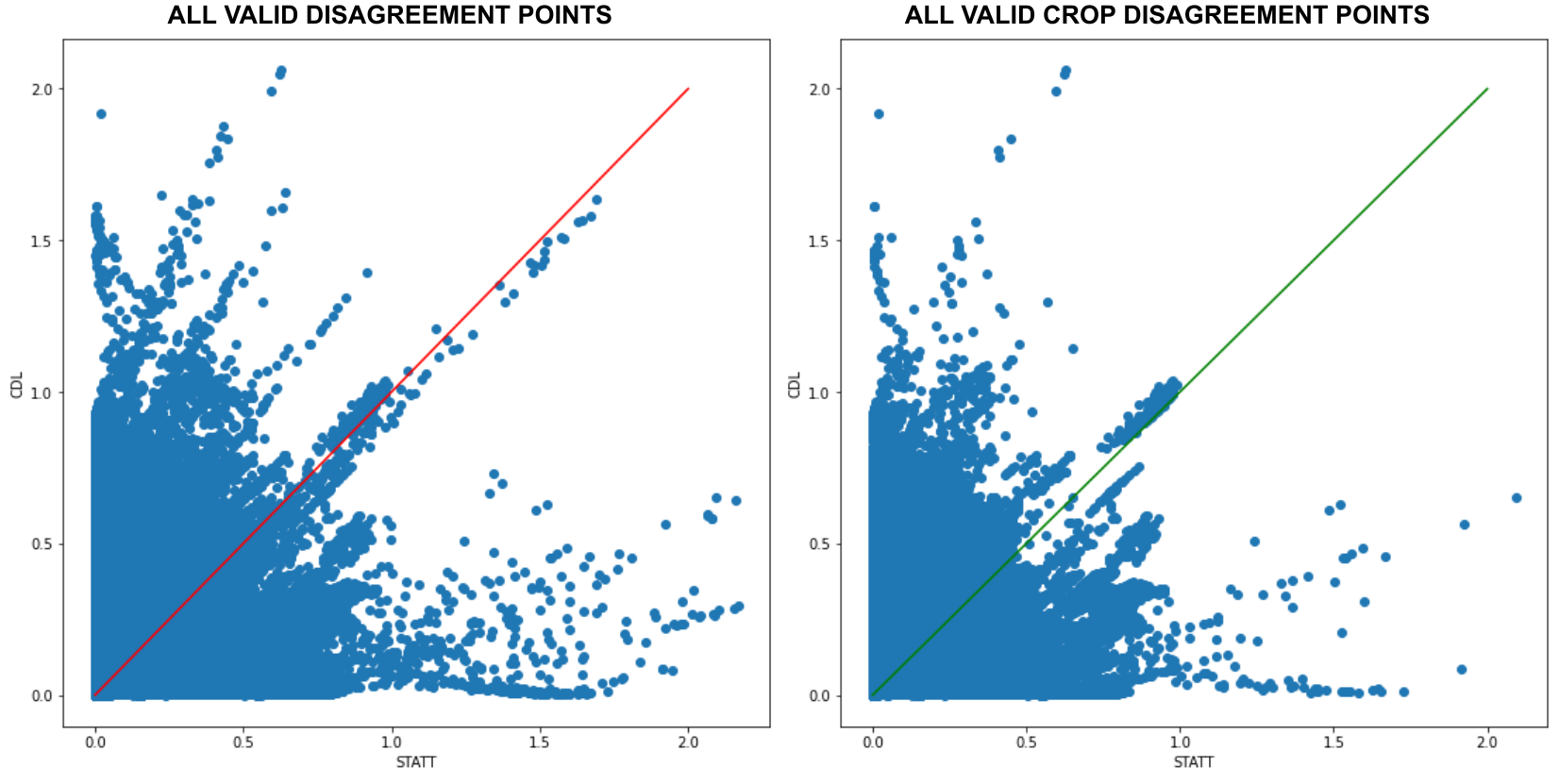}
    \caption{NMSE Scatterplot for all disagreement points as well as crop disagreement points}
    \label{fig:nmsescatterplot}
\end{figure*}

In this section we present a scatter plot based method to globally show improvement over CDL overall and not class wise. As mentioned before, for each point of disagreement CDL predicts a class, which we call CDL class, and STATT predicts another class, which we call STATT class. Now as mentioned in the previous section, we create the characteristic NDVI series for each class using the agreement pixels for each class. Now for each disagreement pixel, we compare the Normalised Mean Square Error and plot in the form of a scatterplot shown in the left of Fig. \ref{fig:nmsescatterplot}. The right scatterplot represents a subset of these disagreement points where at least one of the strategies has predicted a crop class. 

Since the distinction between the methods is not exactly clear from the scatterplots, the statistics for both the plots are present in \ref{Tab:nmsescatterplot}. From the table we can see that when we look at all the pixels overall STATT is closer to the respective characteristic NDVI class series in 1672366 pixels when compared to where CDL is closer,i.e 1009515 pixels. However, when we look at just the crop confusion pixels we can see STATT is better in almost double the amount of pixels that CDL is better in, thus showing global superiority of STATT when compared to CDL.

\begin{table*}[t!]
    \begin{tabular}{lcccccc}
        \hline
        \multicolumn{1}{c}{Class} & \multicolumn{1}{c}{Nmse error threshold} & \multicolumn{1}{c}{$Count_{agreement}$ threshold} & \multicolumn{1}{c}{F1 CDL} & \multicolumn{1}{c}{F1 STATT} & \multicolumn{1}{c}{F1 USDA(rep)} & \multicolumn{1}{c}{\% pixels} \\  \hline
        Almonds                   & 0.5                                      & 100                                       & 0.9513                            & 0.9575                             & 0.9516                       & 0.4104                        \\
        Rice                      & 0.2                                      & 100                                       & 0.9615                            & 0.9845                             & 0.9805                       & 0.4616                        \\
        Alfalfa                & 0.3                                      & 100                                       & 0.8185                            & 0.8612                             & 0.8812                       & 0.3162                        \\
        Pistachio                 & 0.3                                      & 100                                       & 0.8012                            & 0.9150                             & 0.9632                       & 0.2438                        \\
        Grapes                    & 0.3                                      & 100                                       & 0.8945                            & 0.8804                             & 0.9830                       & 0.4293                        \\
        Fallow and Idle         & 0.3                                      & 100                                       & 0.7229                            & 0.8122                             & 0.7183                       & 0.5164                        \\
        Cotton                    & 0.2                                      & 100                                       & 0.9532                            & 0.9464                             & 0.9314                       & 0.4212                        \\
        Walnut                    & 0.2                                      & 100                                       & 0.8859                            & 0.8930                             & 0.9514                       & 0.5281                        \\
        Tomatoes                  & 0.2                                      & 100                                       & 0.8829                            & 0.9650                             & 0.8764                       & 0.3469                        \\
        Winter Wheat             & 0.2                                      & 100                                       & 0.8340                            & 0.8976                             & 0.7000                       & 0.3003                        \\
        Corn                      & 0.2                                      & 100                                       & 0.7734                            & 0.8613                             & 0.7530                       & 0.4092                        \\
        Citrus                    & 0.3                                      & 100                                       & 0.8407                            & 0.8678                             & 0.7583                       & 0.2891                        \\
        Sunflower                 & 0.2                                      & 100                                       & 0.9703                            & 0.8630                             & 0.7956                       & 0.2856                        \\
        Onions                    & 0.1                                      & 100                                       & 0.8064                            & 0.8274                             & 0.7553                       & 0.2205                        \\
        Olives                    & 0.2                                      & 100                                       & 0.8816                            & 0.9663                             & 0.9255                       & 0.2298                        \\
        Dry Beans                 & 0.3                                      & 100                                       & 0.9747                            & 0.9929                             & 0.6503                       & 0.3472                        \\
        Pomegranates              & 0.2                                      & 100                                       & 0.7143                            & 0.7796                             & 0.9448                       & 0.2401                        \\
        Garlic                    & 0.2                                      & 100                                       & 0.9191                            & 0.9572                             & 0.8242                       & 0.3481                        \\
        Cherries                  & 0.4                                      & 100                                       & 0.9005                            & 0.9240                             & 0.9521                       & 0.2533                        \\
        Safflower                 & 0.1                                      & 100                                       & 0.8137                            & 0.8083                             & 0.6899                       & 0.4173                        \\ \hline
        AVG:                      & \multicolumn{1}{l}{}                     & \multicolumn{1}{l}{}                      & 0.8650                            & 0.8980                             & 0.8493                       & \multicolumn{1}{l}{}         \\ \hline
    \end{tabular}
    \caption{NMSE analysis table}
    \label{Tab:nmsetableResults}
\end{table*}

\end{document}